\documentclass[10pt,twocolumn,letterpaper]{article}

\usepackage{cvpr}              

\usepackage[dvipsnames, table]{xcolor}

\definecolor{cvprblue}{rgb}{0.21,0.49,0.74}
\usepackage[pagebackref,breaklinks,colorlinks,citecolor=cvprblue]{hyperref}
\usepackage{times}
\usepackage{epsfig}
\usepackage{graphicx}
\usepackage{dblfloatfix}
\usepackage{amsmath}
\usepackage{amssymb}
\usepackage{booktabs}
\usepackage[T1]{fontenc}
\usepackage{graphics}
\usepackage[capitalize]{cleveref}
\usepackage{booktabs}
\usepackage{makecell}

\usepackage[accsupp]{axessibility}

\def\paperID{CVSPORTS_29} 
\def\confName{CVPR}
\def\confYear{2024}

\title{TeamTrack: \\ A Dataset for Multi-Sport Multi-Object Tracking in Full-pitch Videos}

\author{
    Atom Scott$^{1,4,\ast}$, 
    Ikuma Uchida$^{2,4}$, 
    Ning Ding$^{1}$, 
    Rikuhei Umemoto$^{1}$, 
    Rory Bunker$^{1}$, 
    Ren Kobayashi$^{1}$, \\
    Takeshi Koyama$^{3}$, 
    Masaki Onishi$^{4}$, 
    Yoshinari Kameda$^{2}$, 
    Keisuke Fujii$^{1}$ 
    \\
    [2mm]
    $^1$Nagoya University, 
    $^2$University of Tsukuba, 
    $^3$Tokai University, 
    $^4$AIST
    \\
    \normalsize{$^\ast$Corresponding Author} 
    \\
    \normalsize{
    \url{https://atomscott.github.io/TeamTrack/}
    }
}

\newcommand{\myparagraph}[1]{{\vspace{.5em} \noindent \bf #1}}

\begin{document}
\twocolumn[{%
\renewcommand\twocolumn[1][]{#1}%
\maketitle
\begin{center}
    \centering
    \captionsetup{type=figure}
    \includegraphics[width=0.8\textwidth]{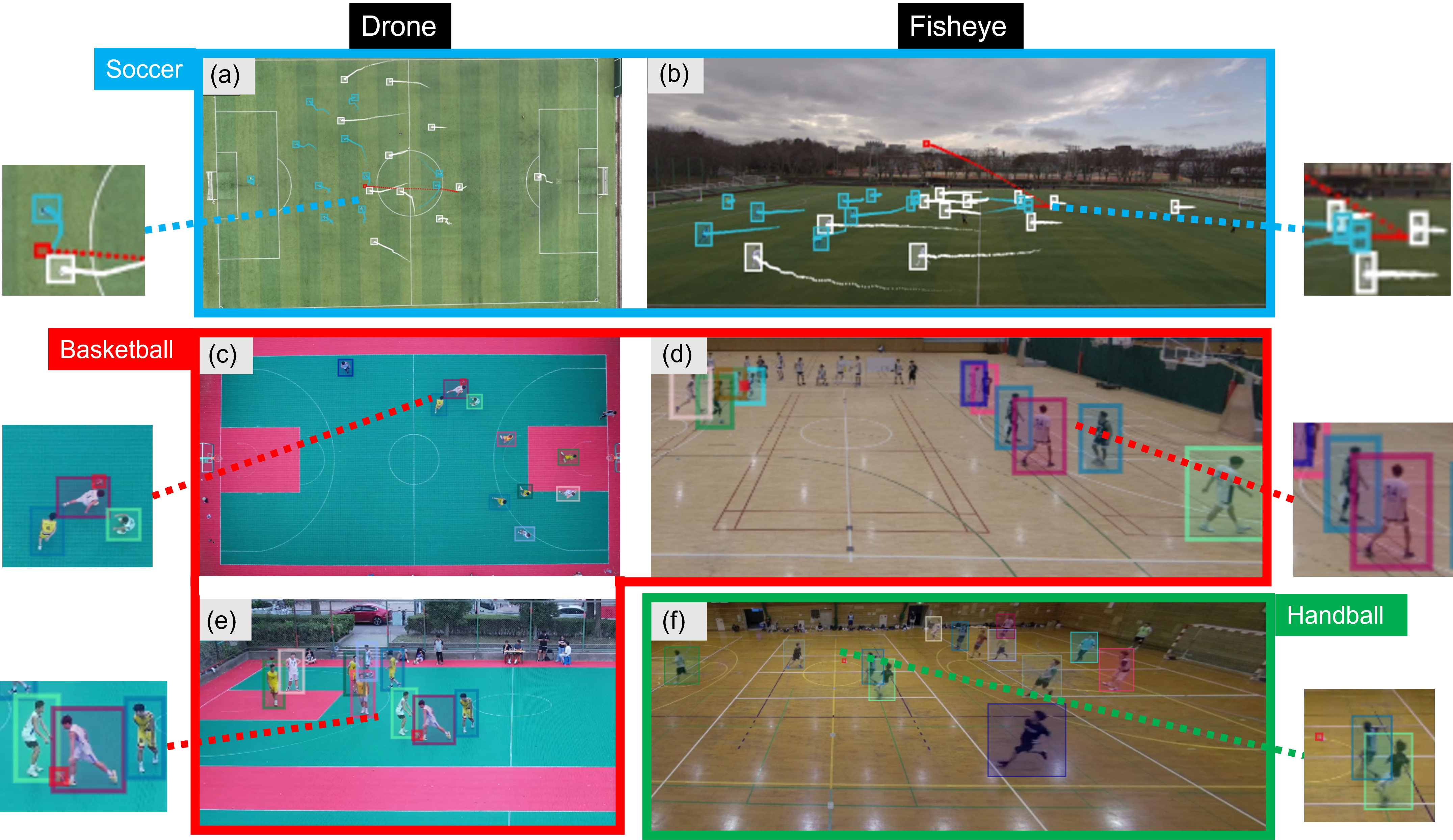}
    \captionof{figure}{Example of our TeamTrack dataset. We provide (a) top-view and (b) side-view in soccer, (c) top-view and, (d)/(e) two side-view videos in basketball, and (f) a single side-view video in handball.
    }
\end{center}%
}]
\begin{abstract}
Multi-object tracking (MOT) is a critical and challenging task in computer vision, particularly in situations involving objects with similar appearances but diverse movements, as seen in team sports. Current methods, largely reliant on object detection and appearance, often fail to track targets in such complex scenarios accurately. This limitation is further exacerbated by the lack of comprehensive and diverse datasets covering the full view of sports pitches. Addressing these issues, we introduce TeamTrack, a pioneering benchmark dataset specifically designed for MOT in sports. TeamTrack is an extensive collection of full-pitch video data from various sports, including soccer, basketball, and handball. Furthermore, we perform a comprehensive analysis and benchmarking effort to underscore TeamTrack's utility and potential impact. Our work signifies a crucial step forward, promising to elevate the precision and effectiveness of MOT in complex, dynamic settings such as team sports. The dataset, project code and competition is released at: \url{https://atomscott.github.io/TeamTrack/}.
\end{abstract}    
\section{Introduction}

\label{sec:intro}

Multi-object tracking (MOT) is a key task in computer vision, critical for many practical applications such as autonomous driving and human movement analysis \cite{rangesh2019no,sun2022dancetrack,milan2016mot16,scott2022soccertrack}. MOT involves localizing and associating objects over time while dealing with unsolved challenges like occlusion, mis-detections, and ID switches. Current methods typically employ detection followed by re-identification (ReID), relying on appearance-based features for temporal association \cite{wojke2017simple,zhou2019omni}. However, these methods falter when confronted with objects of similar appearances performing complex movements, which are common in team sports.

In sports analytics, the use of advanced metrics has increased, and the need for fine-grained tracking data has become increasingly evident \cite{spearman2018beyond,decroos2019actions,teranishi2022evaluation}. Although sensor-based systems, such as the global navigation satellite system and local positioning systems, have emerged as alternatives, these systems have constraints related to availability, budget, and feasibility \cite{ pettersen2014soccer,hennessy2018current}. 
Additionally, gaining information on opposing teams can be problematic. As a result, MOT remains a desirable solution, providing a non-intrusive and affordable method for acquiring data.

In light of these considerations, we introduce the TeamTrack Dataset, a specialized resource developed to address the unique challenges posed by team sports tracking. Unlike existing datasets, which predominantly feature pedestrian movements or are derived from broadcast footage with limited field coverage, TeamTrack offers an expansive view of the game by incorporating over 200,000 frames and 4 million bounding boxes across football, handball, and basketball, marking it as the largest dataset of its kind in terms of volume and scope, as shown in Table~\ref{tab:prior}. This extensive compilation provides a panoramic view of game scenarios to capture dense player formations, swift motion variations, and frequent occlusions, thereby setting a new benchmark for MOT research in sports.

The primary contributions of this paper are three-fold: (1) We create a new dataset\footnote{Available on Kaggle ( \url{kaggle.com/datasets/atomscott/teamtrack}) or Google drive (\url{https://bit.ly/3CMN2hP}).}, comprising of an unprecedented volume of high-resolution, full-pitch video data from diverse team sports, and multiple viewpoints. (2) A detailed exploration of the dataset’s development process, its strategic focus, and an analysis highlighting its characteristics, which promise significant advancements in sports analytics and broader MOT applications. (3) We perform comprehensive evaluations of object detection, trajectory forecasting, and multiple object tracking tasks, setting benchmarks for future research and development.
\section{Related Work}
\begin{table}[!ht]\centering
\rowcolors{2}{gray!25}{white}
\begin{tabular}{lrrc}\toprule
Dataset & Frames & BBoxes & Domain \\\midrule
MOT16~\cite{milan2016mot16} &11,235 &292,733 &Pedestrians \\
MOT20~\cite{dendorfer2020mot20} &13,410 &1,652,040 & Pedestrians \\
KITTI-T~\cite{geiger2015kitti} &10,870 &65,213 & \makecell{Autonomous \\Driving} \\
DanceTrack~\cite{sun2022dancetrack} &105,855 &- &Dance \\
SSET~\cite{feng2020sset} &12,000 &12,000 &Soccer \\
SN-Tracking~\cite{cioppa2022soccernet} &225,375 &3,645,661 &Soccer \\
SportsMOT~\cite{cui2023sportsmot} &150,379 & 1,629,490 &\makecell{Soccer \\ Basketball \\ Volleyball} \\
SoccerTrack\cite{scott2022soccertrack} &82,800 &2,484,000 &Soccer \\\midrule
\makecell{\textbf{TeamTrack}\\\textbf{(ours)}} & \textbf{279,900} &\textbf{4,374,900} &\makecell{\textbf{Soccer} \\ \textbf{Basketball} \\ \textbf{Handball}} \\
\bottomrule
\end{tabular}
\vspace{-10pt}
\caption{Comparative overview of MOT datasets, showcasing TeamTrack with the highest number of frames and bounding boxes.}\label{tab:prior}
\vspace{-15pt}
\end{table}

\label{sec:related}
\vspace{-3pt}
\noindent \textbf{Multi-Object Tracking.}
The typical approach to MOT algorithms follows the tracking-by-detection paradigm. This procedure starts with an object detection using models like RetinaNet \cite{lin2017focal}, CenterNet \cite{duan2019centernet}, or YOLO \cite{redmon2016you}, followed by association via feature extraction, typically achieved by using CNNs \cite{he2016deep,zhou2019omni} and recently ViT models \cite{He2021TransReIDTO,Shen_2022_CVPR}. DeepSORT \cite{wojke2017simple} exemplifies this approach, combining Kalman motion states and deep appearance descriptors for robust tracking. However, these methods often struggle to distinguish and track objects with similar appearances.

End-to-end tracking methodologies, on the other hand, handle object detection and tracking concurrently, offering potential improvements in performance. Models like Tracktor \cite{bergmann2019tracking} leverage frame redundancy to eliminate separate data association, while Neural Solver \cite{braso2020learning} and DeepMOT \cite{xu2020train} deploy neural and Siamese networks, respectively, to address tracking. Recent developments have seen the introduction of Transformer architectures into tracking, such as in DETR \cite{carion2020end}, formulating object detection as a set prediction problem. This paradigm shift has been adopted by models like Trackformer \cite{meinhardt2022trackformer} and TransTrack \cite{sun2020transtrack}. More advanced models, like MOTR \cite{zeng2022motr} and its improved version MOTRv2 \cite{zhang2023motrv2}, extend this framework by adding a query-interaction module to enhance tracking performance. 

\noindent \textbf{Tracking in Sports.}
In sports tracking, a survey \cite{manafifard2017survey} summarized various approaches
using a combination of background subtraction, triangulation from multiple cameras, and Kalman filters to track player movements on the pitch \cite{iwase2004parallel}. Additionally, research exists which addresses this issue by representing player positions as nodes on a graph, and trajectories as edges between nodes \cite{figueroa2006tracking, sullivan2006tracking}. Sullivan et al. proposed a Bayesian framework for linking player IDs \cite{sullivan2009multi}. Furthermore, Lu et al. presented a learning approach based on handcrafted visual features and a Kalman filter to identify and track players within videos \cite{zhang2008collaborate}.

Recent methods have largely focused on using deep learning techniques for player tracking. For instance, Hurault et al. proposed a method of detecting and tracking soccer players using a self-supervised learning method, by transfer learning from an object detection model trained on generic objects \cite{hurault2020self}. Theagarajan et al. identified the player holding the ball using a YOLOv2 network detector and a DeepSORT tracker \cite{theagarajan2018soccer}. Maglo et al. utilized human annotations collected in a semi-interactive system \cite{maglo2022efficient}. Wang et al. proposed improving tracking in various sports scenes by introducing a three-stage matching process to solve motion blur and body overlap \cite{wang2022sportstrack}. 

\noindent \textbf{Multi-Object Tracking Datasets.}
Development of MOT algorithms depends on the availability of annotated datasets. Early examples are the MOT Challenge dataset \cite{milan2016mot16} and the KITTI Tracking Benchmark \cite{geiger2015kitti}, focusing on a range of scenarios from crowded urban environments to autonomous driving. Later datasets like UA-DETRAC \cite{wen2020ua}, PETS \cite{patino2016pets}, and DanceTrack \cite{sun2022dancetrack} further contributed to the field, annotated with bounding boxes and tracking IDs to evaluate algorithmic tracking accuracy.

Tracking in sports presents unique challenges, such as difficulty in differentiating players on the same team and handling occlusion scenarios. 
While datasets like SoccerNet-Tracking \cite{cioppa2022soccernet} and SoccerTrack \cite{scott2022soccertrack} provide valuable resources, their focus is often limited to specific sports or partial field views. Furthermore, the use of broadcast footage in datasets such as SportsMOT \cite{cui2023sportsmot} complicates the direct application of tracking algorithms, necessitating additional processing for image registration, handling zoom, and scene transitions.

The landscape of MOT datasets shows a notable lack of resources tailored for the nuanced challenges of sports tracking. Addressing this, TeamTrack introduces a high-resolution, full-pitch dataset across soccer, basketball, and handball, providing over 279,900 frames and 4,374,900 bounding boxes—the largest in its field. It supports the development of algorithms for complex team sports dynamics. With sports diversity and two perspectives—side and top views—TeamTrack offers comprehensive playfield insights, fostering the development of more sophisticated MOT methods. This dataset aims to drive research forward, making high-quality tracking data accessible and encouraging the creation of robust algorithms to navigate sports tracking's unique challenges.

\begin{table*}[!ht]
\centering
\caption{Camera and video details of the multi-object tracking dataset for various sports.}
\label{tab:data}
\resizebox{\textwidth}{!}{%
\begin{tabular}{@{}lllllrr@{}}
\toprule
Sport & Perspective & Location & Device & Resolution & Minutes & Bounding Box Count \\ 
\midrule
Soccer & Side (Fisheye) & University of Tsukuba, Outdoor & Z CAM E2-F8 & 8K & 30 & 1,242,000 \\
Soccer & Top (Drone) & University of Tsukuba, Outdoor & DJI Mavic 3 & 4K & 30 & 1,242,000 \\
Basketball & Side (Fisheye) & Tokai University, Indoor & Z CAM E2-F6 & 6K & 17.5 & 346,500 \\
Basketball & Top (Drone) & Wuhu Institute of Technology, Outdoor & DJI Mavic 3 & 4K & 24 & 475,200 \\
Basketball & Side (Drone) & Wuhu Institute of Technology, Outdoor & DJI Mavic 3 & 4K & 24 & 475,200 \\
Handball & Side (Fisheye) & Nagoya University, Indoor & Z CAM E2-F6 & 6K & 30 & 594,000 \\ 
\midrule
Total & & & & & 155.5 & 4,374,900 \\ 
\bottomrule
\end{tabular}%
}
\vspace{-10pt}
\end{table*}
\label{sec:dataset}

\vspace{-5pt}

\section{TeamTrack Dataset}
\vspace{-3pt}
In this section, we present TeamTrack, a novel MOT dataset comprising over 150 minutes of high-resolution video from multiple team sports. The main characteristics of TeamTrack are the following:

\noindent\textbf{Large-scale}: TeamTrack introduces over 4 million annotated bounding boxes across various tracklets, making it one of the largest datasets of its kind, as compared to others listed in Table \ref{tab:data}.

\noindent\textbf{Multiple Sports}: TeamTrack includes matches from three team sports; soccer, basketball, and handball.

\noindent\textbf{Similar Appearance and Dynamic Movement}: The dataset features targets with similar appearance, dynamic movements, and frequent occlusions, offering a robust challenge for tracking algorithms.

\noindent\textbf{Full Pitch Multi Angle View}: Videos are recorded from two angles, top view via drones and side view with fisheye lenses, both covering the entire playing field.

Beyond its significant size and the inherent challenges it presents, TeamTrack's unique value lies in its multi-view and full-pitch capture settings. These features enable experimentation with multi-view tracking and the use of prior information (e.g., player count, pitch dimensions) not feasible with traditional broadcast videos. Additionally, we hope the adversarial-cooperative nature of team sports can also be studied to further refine tracking techniques.


In the following subsections, we describe the data collection process, annotation and labeling, appearance similarity, and motion patterns observed in the TeamTrack dataset, following the methodology described in DanceTrack \cite{sun2022dancetrack}. Our dataset was compared to DanceTrack, MOT17, and MOT20—datasets characterized by limited camera movements. We did not compare with broadcast video datasets, such as SoccerNet-Tracking \cite{cioppa2022soccernet} and SportsMOT \cite{cui2023sportsmot}, due to their extensive camera movements, which hinder fair comparison.

\subsection{Data Collection}
The data collection process captured video footage across three team sports: soccer, basketball, and handball, recorded at various university venues, encompassing both indoor and outdoor environments. The side and top views were recorded using fisheye and drone cameras, respectively. To correct for the distortion caused by the fisheye lens, we applied Zhang's calibration method. Further information on the devices used and their resolution specifications is detailed in Table \ref{tab:data}.

\vspace{-3pt}
\subsection{Annotation and Labeling}

\begin{figure}[!t]
\centering
\includegraphics[width=0.48\textwidth]{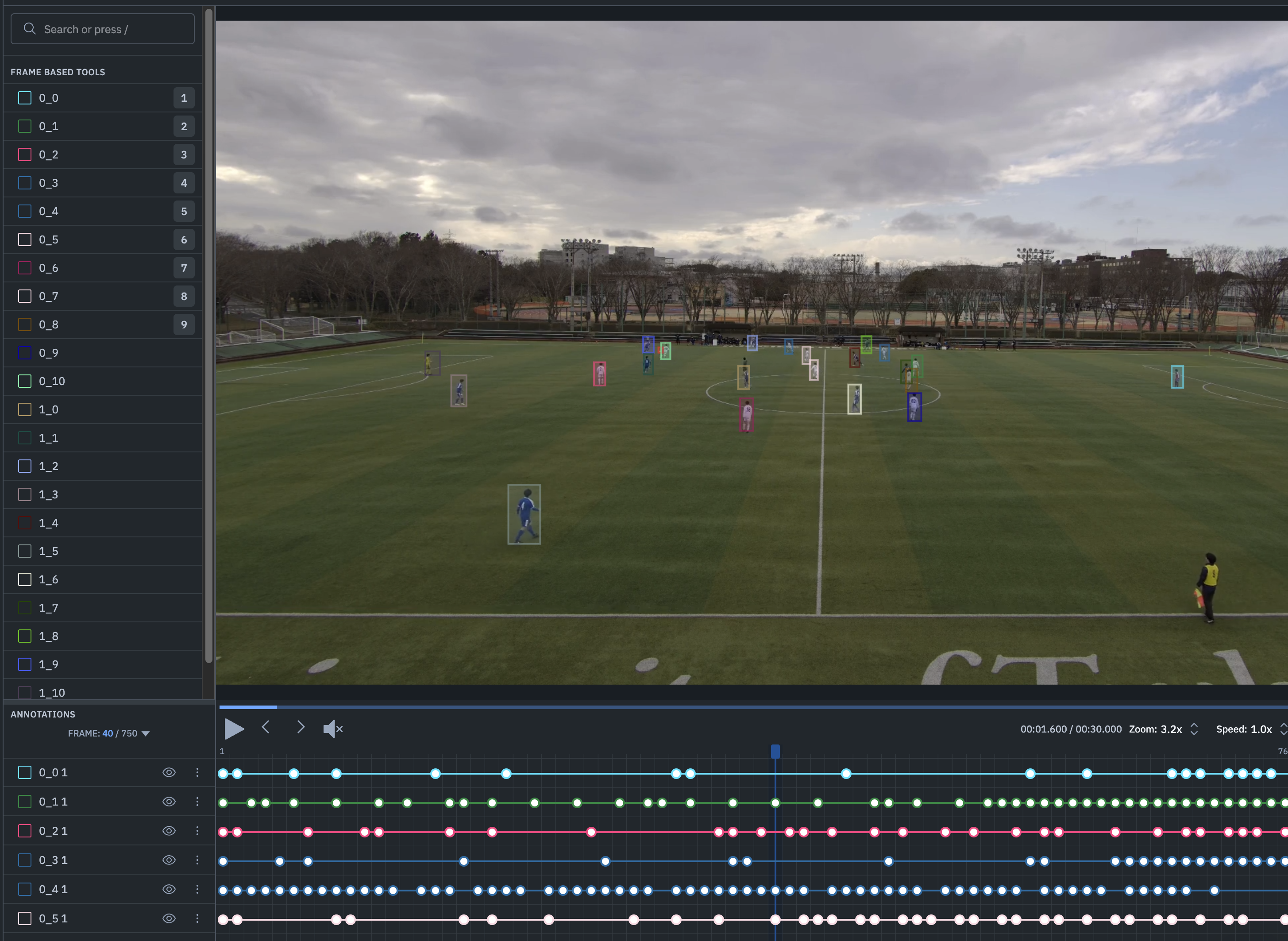}
\caption{Screenshot of Labelbox during the annotation process, illustrating manual adjustments to automated tracking and interpolation annotations.}
\label{fig:annotation_process}
\end{figure}

The annotation process involved a series of strategic decisions and tool evaluations to balance efficiency, accuracy, and cost. Initially, we utilized the open-source tool CVAT but later transitioned to Labelbox to alleviate the burdens associated with working on a local server and to streamline workflows more effectively. Although Labelbox did also offer automated tracking, the practical gains in speed were limited due to inference time not being fast enough.

\noindent\textbf{Interpolation and Manual Annotation:} Both CVAT andLabelbox support interpolation, significantly reducing the need to annotate every frame. This feature enabled focused manual annotation on approximately every 1 to 50 frames, as depicted in the timeline of Fig.~\ref{fig:annotation_process}, especially in scenes experiencing dynamic changes. While this method streamlined the workflow, it still required significant manual input to ensure accuracy.

\noindent\textbf{Challenges with Semi-Automatic Detection:} We experimented with semi-automatic detection using a pretrained object detector for initial annotations. However, the time saved was often offset by the need to correct false positives/negatives and manage ID switches. Labelbox's lack of track merging capabilities further complicated these adjustments, as illustrated by the comparative analysis of pretrained versus fine-tuned object detectors' accuracy in Table.~\ref{tab:object_detection_accuracy}, indicating the necessity for model fine-tuning.

\noindent\textbf{Exploration of Other Tools:} In search of more efficient solutions, we considered advanced annotation tools like V7 and Encord, known for their sophisticated track merging and smart interpolation capabilities. However, the high initial limited their adoption. 

\noindent\textbf{Annotation Effort:} Ultimately, the annotation process required over 600 person-hours, translating to roughly 2 hours for every 30 seconds of video. This significant investment of time and resources highlighted a critical challenge: the pressing need for more efficient annotation tools to facilitate dataset expansion and enhance the scalability of sports analytics research. We hope these insights offer valuable perspectives to the research community on balancing the capabilities, efficiency, and costs of annotation tools.

\noindent\textbf{Dataset Metadata:} The TeamTrack dataset provides persistent player IDs throughout the entirety of a match, although it does not include jersey numbers. Annotations are available in two formats: the SportsLabKit format\footnote{\noindent\url{https://github.com/AtomScott/SportsLabKit}.}, which encompasses team affiliations to enrich analyses of team dynamics and interactions, and the standard MOTChallenge format\footnote{\noindent\url{https://motchallenge.net/instructions/}.}. The latter is especially useful for general tracking studies and conforms to conventional evaluation methodologies.

\subsection{Appearance Similarity}

    
\begin{figure}[t]
     \centering
     \includegraphics[width=\linewidth]
     {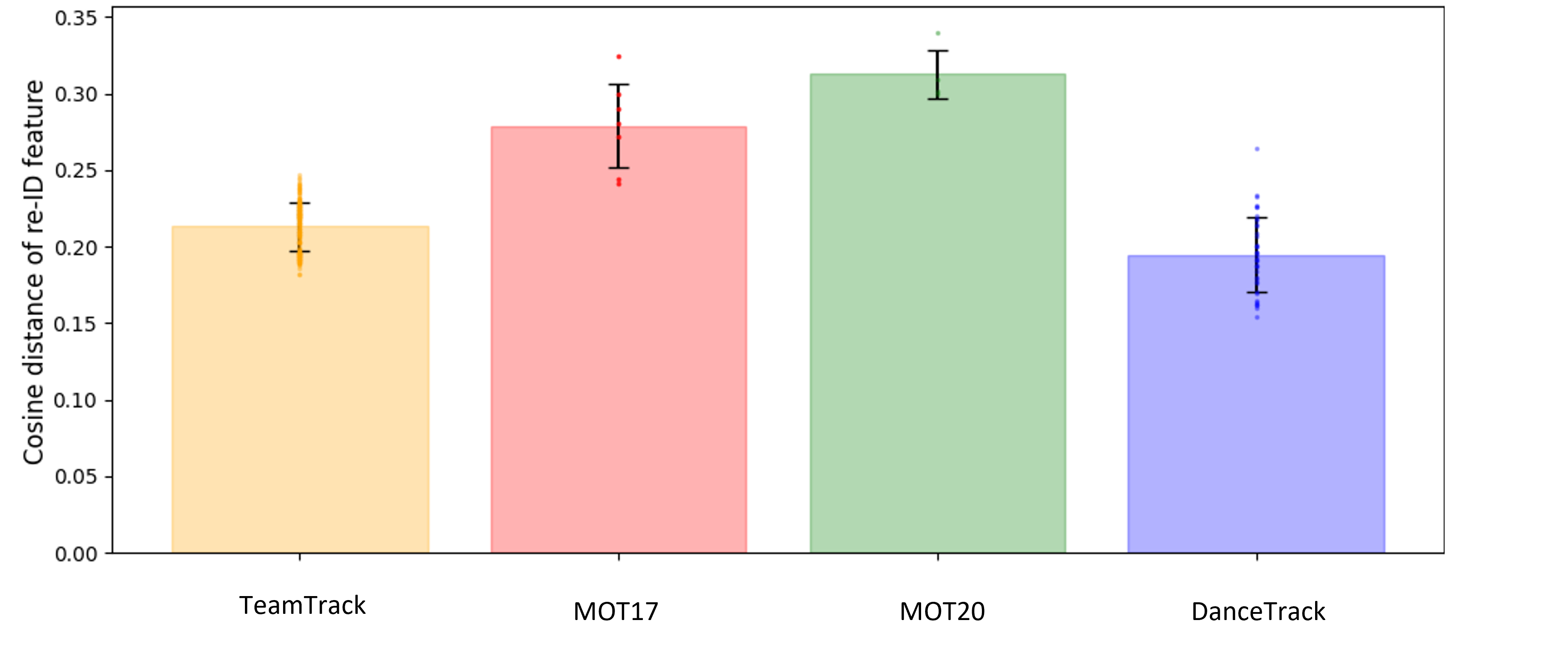}
     \vspace{-7pt}
     \caption{Cosine distance of re-ID features: TeamTrack compared to DanceTrack, MOT17, and MOT20.}
     \label{fig:cos-dist-re-id-feature-4}
     \vspace{-8pt}
 \end{figure}

 \begin{figure}[t]
     \centering
     \includegraphics[width=0.5\textwidth]
     {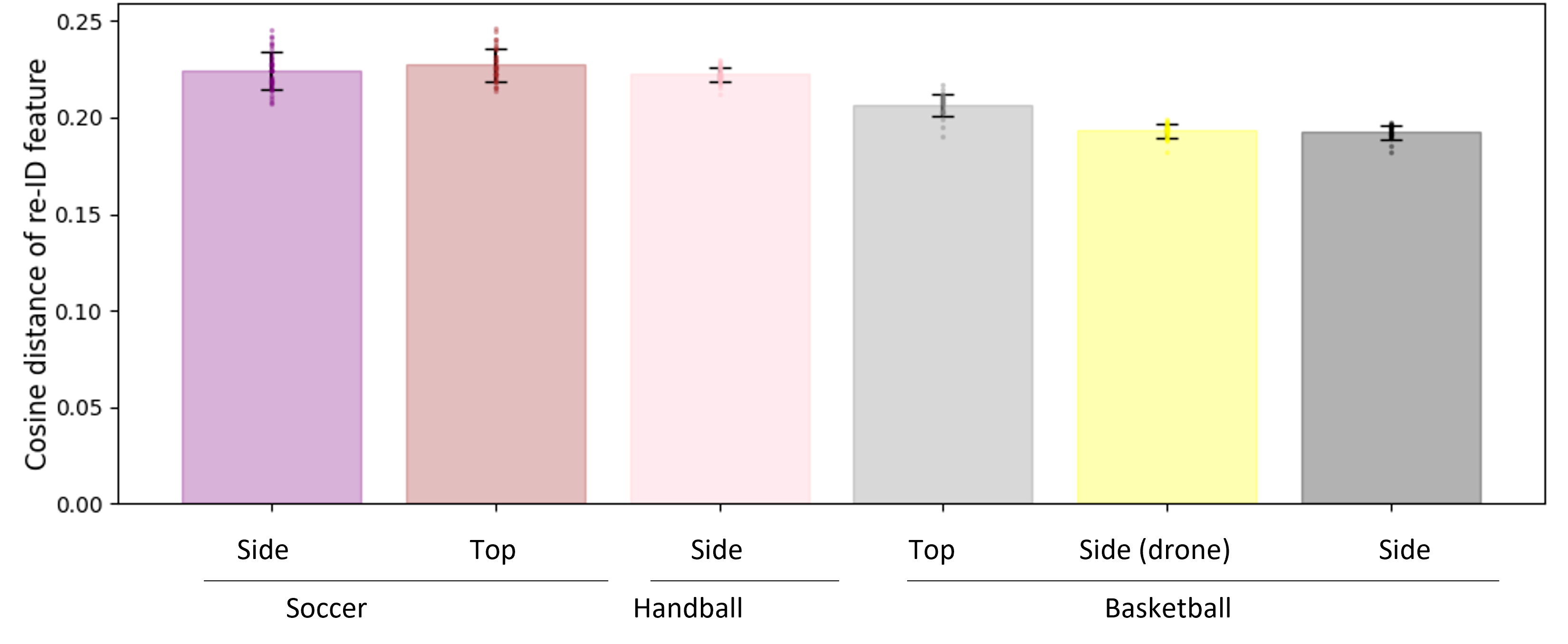}
     \vspace{-15pt}
     \caption{Cosine distance of re-ID features within TeamTrack datasets.}
     \label{fig:cos-dist-re-id-feature-6}
     \vspace{-8pt}
 \end{figure}
\vspace{-3pt}

\vspace{-2pt}
\begin{figure*}[!t]
    \centering
    
    \setlength{\fboxrule}{.5px}
    \setlength{\fboxsep}{0px}
    
    \begin{subfigure}[b]{0.3\textwidth}
        \fbox{\parbox[c][1.1\textwidth][c]{1.1\textwidth}{\centering\includegraphics[width=\textwidth]{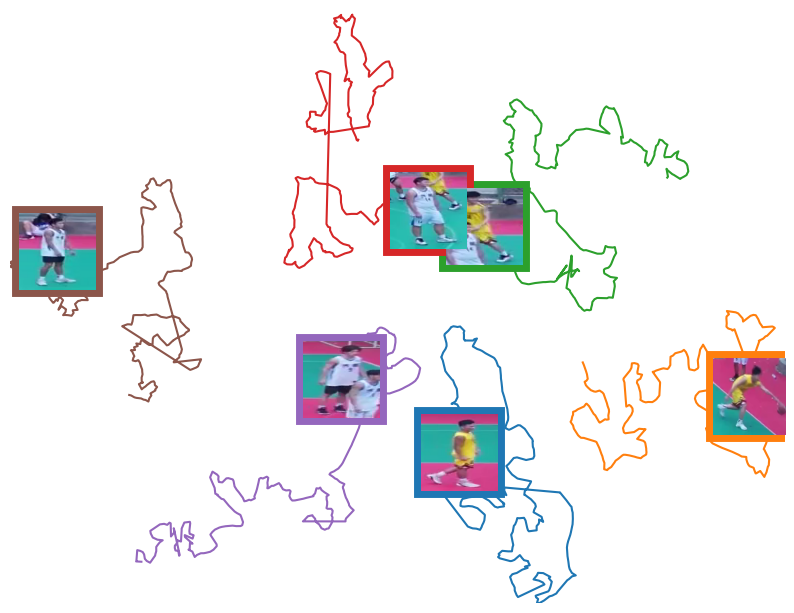}}}
        \caption{Basketball Sideview 1}
    \end{subfigure}
    \hfill
    \begin{subfigure}[b]{0.3\textwidth}
        \fbox{\parbox[c][1.1\textwidth][c]{1.1\textwidth}{\centering\includegraphics[width=\textwidth]{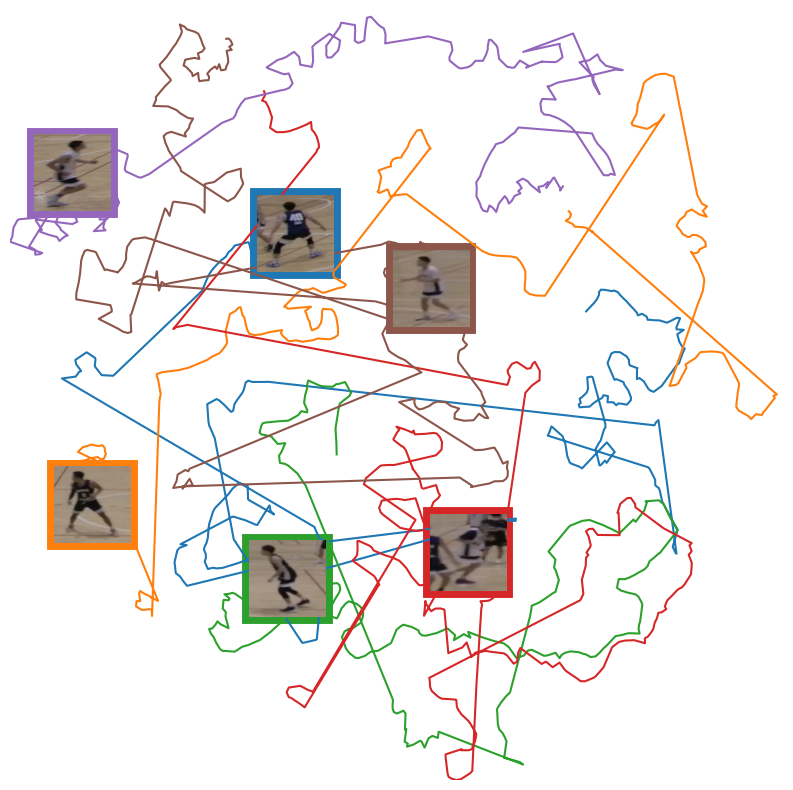}}}
        \caption{Basketball Sideview 2}
    \end{subfigure}
    \hfill
    \begin{subfigure}[b]{0.3\textwidth}
        \fbox{\parbox[c][1.1\textwidth][c]{1.1\textwidth}{\centering\includegraphics[width=\textwidth]{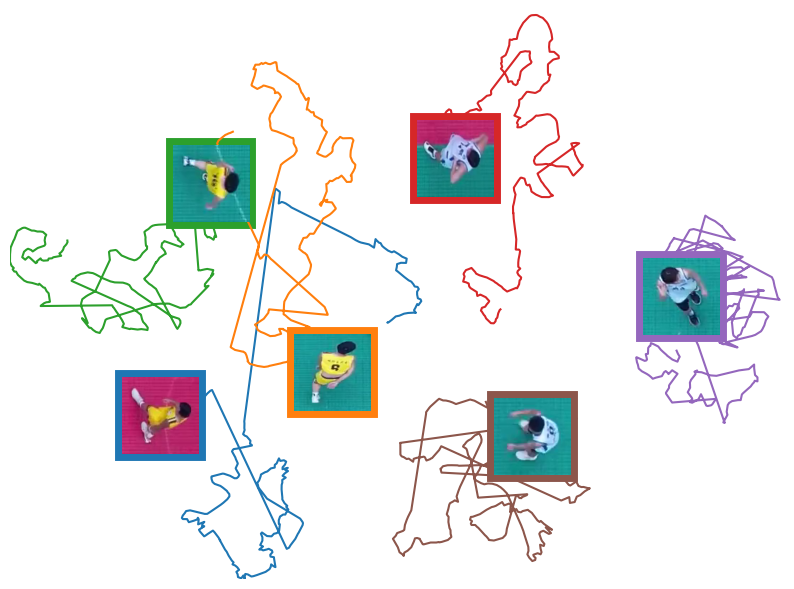}}}
        \caption{Basketball Topview}
    \end{subfigure}
    
    \vspace{1em}
    
    \begin{subfigure}[b]{0.3\textwidth}
        \fbox{\parbox[c][1.1\textwidth][c]{1.1\textwidth}{\centering\includegraphics[width=\textwidth]{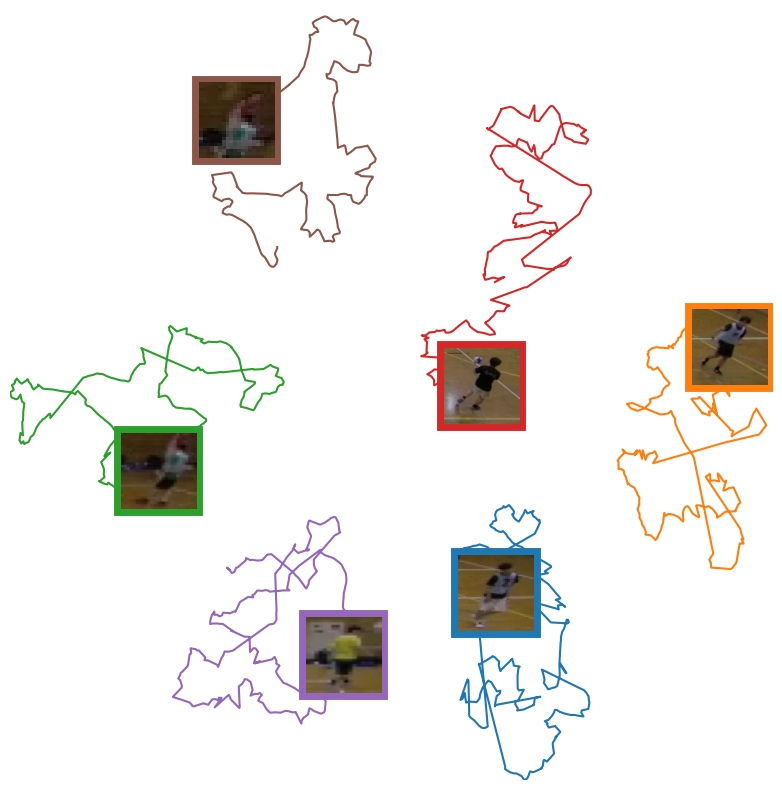}}}
        \caption{Handball Sideview}
    \end{subfigure}
    \hfill
    \begin{subfigure}[b]{0.3\textwidth}
        \fbox{\parbox[c][1.1\textwidth][c]{1.1\textwidth}{\centering\includegraphics[width=\textwidth]{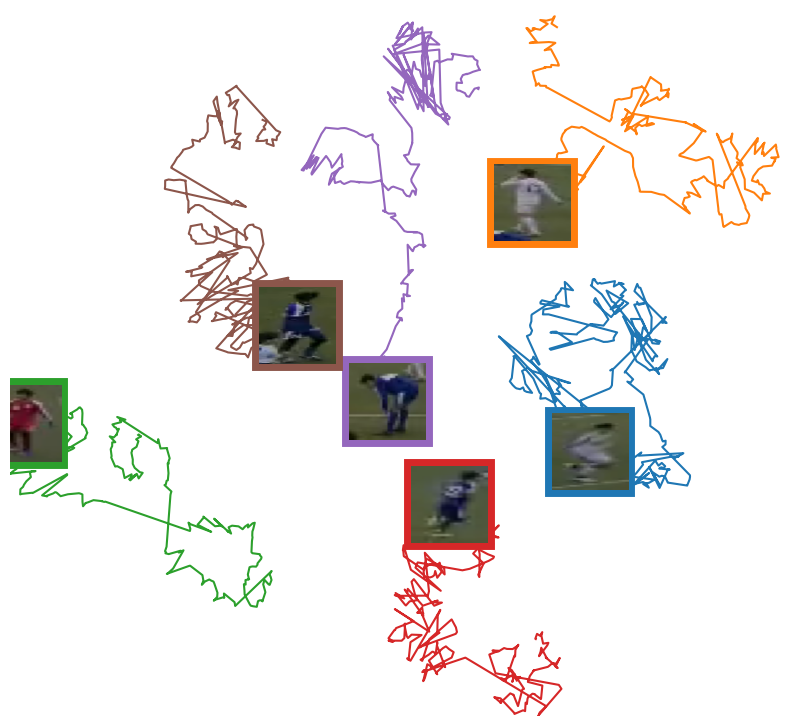}}}
        \caption{Soccer Sideview}
    \end{subfigure}
    \hfill
    \begin{subfigure}[b]{0.3\textwidth}
        \fbox{\parbox[c][1.1\textwidth][c]{1.1\textwidth}{\centering\includegraphics[width=\textwidth]{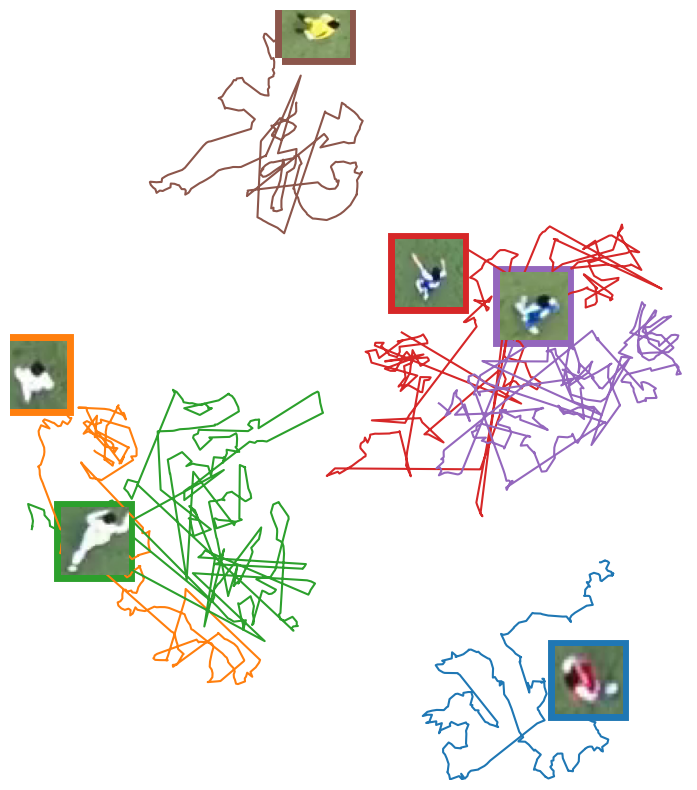}}}
        \caption{Soccer Topview}
    \end{subfigure}
    
    \caption{Visualization of re-ID features from various videos in our TeamTrack dataset using t-SNE. Objects are color-coded for consistent identification. The bounding box from the first frame is superimposed on the corresponding re-ID feature for contextualization.}
    \label{fig:tsne_visualizations}
\end{figure*}

\noindent\textbf{Quantitave analysis of appearance features}: To quantitatively test our hypothesis that the targets in teamtrack have similar appearances, we employ a pre-trained re-ID model \cite{sun2022dancetrack} to extract the appearance features and calculate a similarity metric. The similarity metric is based on the cosine distances between these features across video frames:
\[
V = \frac{1}{T} \sum_{t=1}^{T} \frac{1}{N_t^2} \sum_{i=1}^{N_t} \sum_{j \neq i}^{N_t} \left(1 - \cos < F(B_{ti}), F(B_{tj}) >\right),
\]
where $F(B_{ti})$ and $ F(B_{tj})$ are appearance features, $B$ and $T$ are an object and the number of frames in the video, $N_t$ is the number of objects on the frame $t$ and $<\cdot>$ is the angle between two vectors. As depicted in Figure \ref{fig:cos-dist-re-id-feature-4}, TeamTrack datasets generally demonstrated a higher appearance similarity among players compared to the MOT datasets, with lower mean cosine distances. Figure \ref{fig:cos-dist-re-id-feature-6} shows the comparison within TeamTrack datasets, highlighting that the similarity metric is somewhat consistent across sports.

\noindent\textbf{Visualization of appearance features}: To further understand the visual characteristics of the TeamTrack dataset, we present visualizations of appearance features extracted from several videos. In Figure \ref{fig:tsne_visualizations}, we show t-SNE embeddings of image Re-ID features, extracted from the first 200 frames of each dataset. These visualizations reveal that, although appearance features can be differentiated in certain sports—particularly under varying background conditions and among players from opposing teams—features of players within the same team tend to be highly entangled. It should be noted that as the number of frames increases, the bounding boxes for players are likely to overlap, adding another layer of complexity to the task. This qualitatively demonstrates the high degree of appearance similarity among within the proposed TeamTrack dataset.

\vspace{-3pt}
\subsection{Motion Patterns}
\vspace{-2pt}
TeamTrack dataset includes complex group motion patterns compared to popular pedestrian datasets. To quantitatively confirm this, we computed IoU (Intersection-over-Union) on adjacent frames and frequency of relative position switch as motion patterns metrics in the TeamTrack dataset and compare them with other MOT datasets, as described in DanceTrack \cite{sun2022dancetrack}.

\noindent \textbf{IoU on adjacent frames:} We calculate the average IoU of bounding boxes on adjacent frames for each video. A low IoU implies fast-moving objects or low video frame rates. The averaged IoU on adjacent frames for a video with $N$ objects and $T$ frames is defined as follows:
\[
U = \frac{1}{N(T-1)} \sum_{i=1}^{N} \sum_{t=1}^{T-1} IoU(B_{ti}, B_{t+1i}).
\]

\begin{figure}[t]
    \centering
    \includegraphics[width=0.45\textwidth]{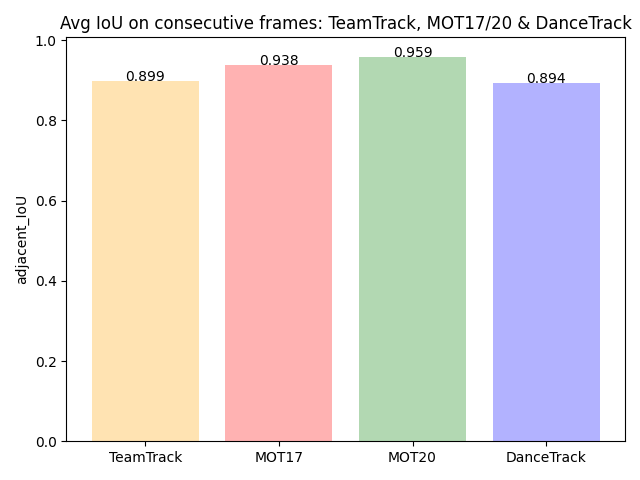}
    \vspace{-7pt}
    \caption{Average IoU on adjacent frames: TeamTrack compared with MOT17, MOT20, and DanceTrack.}
    \label{fig:iou-teamtrack-vs-mot-dancetrack}
    \vspace{-8pt}
\end{figure}

Figure \ref{fig:iou-teamtrack-vs-mot-dancetrack} shows the IoU on adjacent frames in TeamTrack and other datasets.
The results indicate that TeamTrack had a slightly lower average IoU on adjacent frames compared to MOT17/MOT20, and a similar average IoU on adjacent frames to DanceTrack. The lower average IoU on adjacent frames suggests that objects/regions move, change in shape or size, or disappear to a greater extent in TeamTrack and DanceTrack compared to MOT17/MOT20.

\begin{figure}[t]
    \centering
    \includegraphics[width=0.45\textwidth]{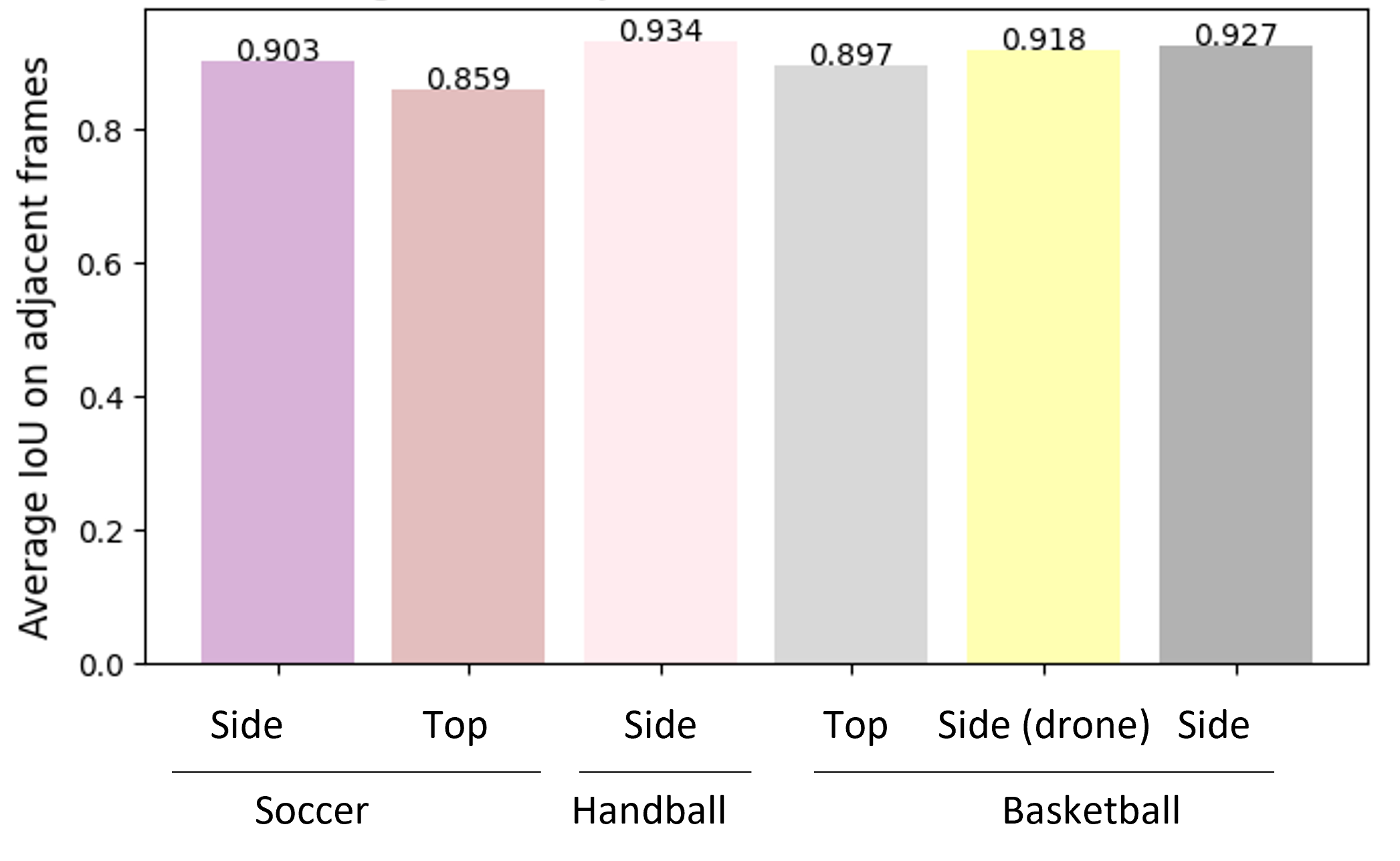}
    \vspace{-7pt}
    \caption{Average IoU on adjacent frames: within TeamTrack datasets.}
    \label{fig:iou-teamtrack}
    \vspace{-8pt}
\end{figure}

 Figure \ref{fig:iou-teamtrack} shows the results among the six TeamTrack datasets. Soccer Top View and Handball Side View had the lowest and highest average IoU on adjacent frames, respectively, indicating that objects maintained their position, size, and shape to a greater extent in handball. 

\begin{figure}[t]
    \centering
    \includegraphics[width=0.45\textwidth]{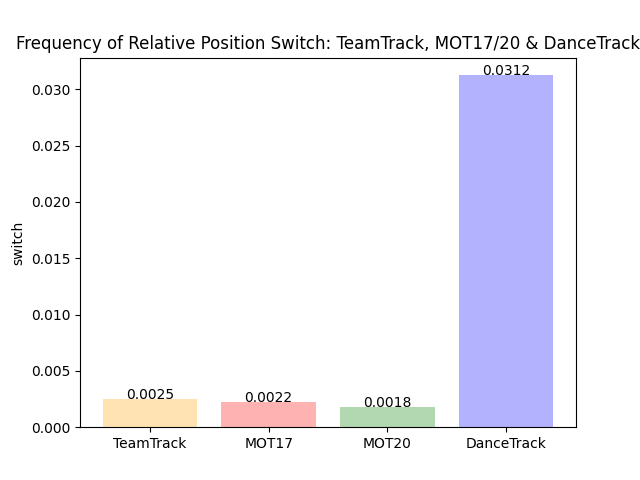}
    \caption{Frequency of relative position switch: TeamTrack compared with MOT17, MOT20, and DanceTrack}
    \label{fig:switch-freq-teamtrack-vs-mot-dancetrack}
\end{figure}

\begin{figure}[t]
    \centering
    \includegraphics[width=0.45\textwidth]{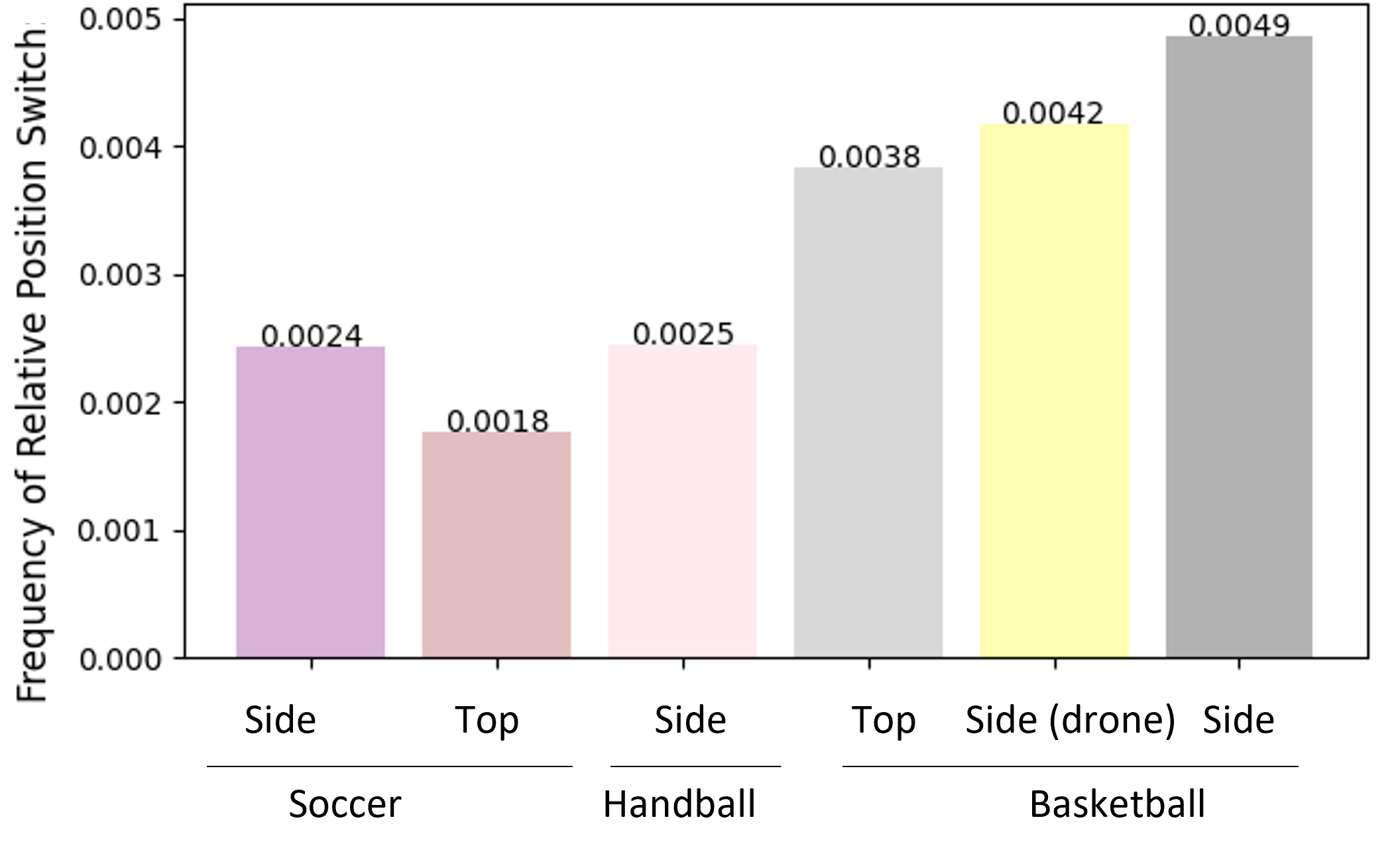}
    \vspace{-7pt}
    \caption{Frequency of relative position switch: within TeamTrack datasets.}
    \vspace{-8pt}
    \label{fig:switch-freq-teamtrack}
\end{figure}

\noindent \textbf{Frequency of relative position switch:} To measure the diversity of object motions, we compute the average frequency of relative position switches for each video based on \cite{sun2022dancetrack}. Frequent position switches can be attributed to highly non-linear motion patterns and result in frequent crossovers and inter-object occlusions. 
The average frequency of relative position switch is defined as follows:
\[
S = \frac{\sum_{i=1}^{N} \sum_{j \neq i}^{N} \sum_{t=1}^{T-1} sw(B_{ti}, B_{tj}, B_{t+1i}, B_{t+1j})}{2N(T - 1)(N - 1)},
\]
where $sw$ is an indicator function, where $sw(\cdot)=1$ if the two objects swap their left-right relative position or top-down relative position on the adjacent frames, $sw(\cdot)=0$ if there is no swap. 

Figure \ref{fig:switch-freq-teamtrack-vs-mot-dancetrack} shows the results in TeamTrack and other datasets, with DanceTrack showing significantly higher frequencies and TeamTrack marginally exceeding those of MOT datasets. This likely stems from the dynamics of team sports, where intense activity often concentrates around the vicinity of the ball or specific events such as corner kicks, potentially leading to an underrepresentation in averaged data. Further analysis within TeamTrack, as illustrated in Figure \ref{fig:switch-freq-teamtrack}, reveals considerable differences among sports. In the context of sports, we interpet the frequency of relative position switch to indicate the activity/intensity level of the match; therefore, in this case, Basketball had a higher level of intensity than Soccer and Handball. 

\vspace{-7pt}
\section{Experiments and Evaluation}
\label{sec:experiments}
\vspace{-3pt}
Our evaluation showcases the utility of the TeamTrack dataset and the distinct challenges it introduces for advancing research in MOT. We concentrate on three pivotal components: object detection (Section \ref{ssec:obj_det}), trajectory forecasting (Section \ref{ssec:trajectory_forecasting}), and finally, multi-object tracking (Section \ref{ssec:multiple_object_tracking}). In each task we followed a conventional data splitting ratio of 70:15:15 for training, validation, and testing sets.

\subsection{Object Detection}
\label{ssec:obj_det}
Object detection is a crucial step for accurately tracking an object. Detection accuracy typically correlates with tracking performance, particularly with methods that employ the tracking-by-detection paradigm.

\noindent\textbf{Setup:}
We fine-tuned the YOLOv8 model \cite{Jocher_YOLO_by_Ultralytics_2023} on each dataset. We adjusted the image size using YOLOv8m, YOLOv8n, and YOLOv8x. The image sizes employed were 512, 1280, and 2560. The training utilized the AdamW optimizer\cite{loshchilov2017decoupled} with an initial learning rate of 0.001, alongside data augmentation techniques (mosaic, flipping, scaling, color jittering) to enhance model robustness. Test-Time Augmentation (TTA) was also implemented.

\noindent\textbf{Metrics:}
We evaluate the performance of the model using the mean Average Precision (mAP) with intersection over union (IoU) thresholds ranging from 0.5 to 0.95 (mAP\textsubscript{50:95}). This metric measures the model's accuracy in terms of both position and size of the predicted bounding boxes. 

\noindent\textbf{Results:} Table~\ref{tab:object_detection_accuracy} showcases YOLOv8's object detection performance across various sports, separated by camera perspectives. The notably lower mAP scores for pre-trained and fine-tuned models in top-view setup can be largely attributed to the smaller appearance of players from high-altitude drone footage. This results in smaller IoU values, making it challenging to achieve high (mAP\textsubscript{50:95}) scores. Nevertheless, fine-tuning significantly improved the mAP scores for all sports and views. The result underscores the importance of fine-tuning for enhancing model performance and highlight the influence of camera perspective on detection accuracy.

\begin{table}[h]
\centering
\caption{COCO mAP\textsubscript{50:95} performance of the object detection model on different sports, split by TV (Top-View) and SV (Side-View) camera perspectives. 
For basketball, SV* denotes the side view with drone camera.}
\begin{tabular}{lccccccc}
\toprule
& \multicolumn{2}{c}{Football} & \multicolumn{3}{c}{Basketball} & \multicolumn{2}{c}{Handball} \\
\cmidrule(lr){2-3} \cmidrule(lr){4-6} \cmidrule(lr){7-8}
YOLOv8 & TV & SV & TV & SV & SV* & \quad SV & \\
\midrule
{\small Pre-trained} & 1.4 & 11.2 & 2.5 & 15.4 &  10.5 & \quad 9.8 & \\
{\small \bf Fine-tuned} & \textbf{23.5} & \textbf{52.7} & \textbf{66.6} & \textbf{68.7} & \textbf{68.6} & \quad \textbf{71.0} & \\
\bottomrule
\end{tabular}
\label{tab:object_detection_accuracy}
\end{table}

\subsection{Trajectory Forecasting}
\label{ssec:trajectory_forecasting}

Trajectory forecasting involves predicting future positions of subjects and can enhance object tracking. For instance, it facilitates the re-identification or continued tracking of objects lost due to occlusions, This is particularly critical in scenarios where appearance cues are unreliable, such as in team sports. We implement both Constant Velocity and LSTM motion models to analyze the movement patterns of individual players within the teamstrack dataset, aiming to grasp the motion characteristics of the TeamTrack dataset.

\noindent\textbf{Setup:}
We adopted the experiment settings from social-LSTM~\cite{alahi2016social}, observing trajectories for 3.2 seconds and predicting the trajectories for the next 4.8 seconds (144 frames). The data was centralized and normalized, and a 5\% overlap sliding window was employed. Data augmentation was carried out by horizontal and vertical flipping on each sequence. Standard procedures, such as teacher forcing and learning rate scheduling, were implemented.

\noindent\textbf{Metrics:}
Our trajectory forecasting model is assessed based on its Root Mean Square Error (RMSE) for one step ahead prediction. RMSE is a standard metric used in trajectory forecasting as it provides an aggregate measure of prediction error. It quantifies how much, on average, our model's predictions deviate from the actual data. A lower RMSE indicates a more accurate model, providing a reliable means of comparing different models or setups.

\noindent\textbf{Results:} Table \ref{tab:trajectory_forecasting_accuracy} showcases the RMSE of both motion models across different sports, measured immediately (1 frame) and a few seconds (144 frames) post-observation. Errors tend to accumalte over time. LSTM generally outperforms, validating the hypothesis that TeamTrack dataset has diverse and non-linear movement. Incorporating team labels and models that capture player interactions, such as graph neural networks, could further improve accuracy, presenting an avenue for future research.

\begin{table}[h]
\centering
\caption{RMSEs of the trajectory forecasting models on different sports, which are averaged over all trajectories.}
\begin{tabular}{lcccccc}
\toprule
Video & \multicolumn{2}{c}{Football} & \multicolumn{2}{c}{Basketball} & \multicolumn{2}{c}{Handball} \\
\cmidrule(lr){2-3} \cmidrule(lr){4-5} \cmidrule(lr){6-7}
Model/Frames & 1 & 144 & 1 & 144 & 1 & 144 \\
\midrule
Constant & 0.04 & 7.99 & 0.06 & 7.68 & 0.04 & 6.08 \\
LSTM & \textbf{0.02} & \textbf{6.50} & \textbf{0.04} & \textbf{5.19} & \textbf{0.02} & \textbf{4.07} \\
\bottomrule
\end{tabular}
\label{tab:trajectory_forecasting_accuracy}
\end{table}


\subsection{Multiple Object Tracking}
\label{ssec:multiple_object_tracking}

In this section, we present a direct evaluation our proposed TeamTrack dataset by applying two state-of-the-art tracking algorithms, ByteTrack and BoT-SORT. Additionally, we compare these results with earlier benchmarks performed on the MOT17 and DanceTrack datasets to highlight the challenges posed by our TeamTrack dataset.

\noindent\textbf{Setup:}
Similar our prior experiments, we split the dataset and proceeded to train YOLOv8 detectors on the training set, followed by hyperparameter optimization using the validation set. For our evaluations, we employed the Ultralytics versions of ByteTrack and BoT-SORT. When comparing to the DanceTrack and MOT17 datasets, we concentrated on ByteTrack alone due to the public availability of results for both datasets, allowing for a direct comparison of performance across all sports within the TeamTrack dataset.

\noindent\textbf{Metrics:} To evaluate the different aspects of tracking, we report results using several commonly used metrics: HOTA, MOTA, IDF1, DetA, and AssA. Each metric uniquely emphasizes aspects of detection (DetA), association performance (AssA, IDF1), or both (HOTA, MOTA), providing a comprehensive evaluation of tracking capabilities.

\noindent\textbf{Results:} The results across different sports and view perspectives, detailed in Table ~\ref{tab:tracking_results}. Both methods showed similar trends. In the Handball SideView, models achieved the highest HOTA, indicating effective tracking capabilities in this setting. The Basketball scenarios, SideView, SideView2 and TopView, revealed contrasting outcomes: the SideView2 posed significant challenges, while the TopView showed better performance metrics, suggesting variations in model efficacy based on the angle of view. In the Soccer datasets, the SideView perspective generally yielded better performance compared to the TopView. Table~\ref{tab:tracking_metrics_over_datasets} shows a comparison of these scores compared to those of other datasets. Although TeamTrack's average scores across all sports do not position it as the most challenging dataset, the variability within specific sports and perspectives underscores its diversity and complexity. This variability marks TeamTrack as a uniquely challenging dataset for developing and evaluating tracking algorithms.

\begin{table}[h]
\centering
\caption{Performance of the various MOT models.}

\subcaptionbox{Soccer SideView}{
\begin{tabular}{lccccc}
\toprule
Method & HOTA & DetA & AssA & MOTA & IDF1 \\
\midrule
BoT-SORT & 58.4 & 62.8 & 54.5 & 84.2 & 73.8 \\
ByteTrack & 59.3 & 64.4 & 54.7 & 86.4 & 74.2 \\
\bottomrule
\end{tabular}
}

\subcaptionbox{Soccer TopView}{
\begin{tabular}{lccccc}
\toprule
Method & HOTA & DetA & AssA & MOTA & IDF1 \\
\midrule
BoT-SORT & 51.9 & 51.1 & 53.3 & 42.7 & 65.7 \\
ByteTrack & 53.7 & 51.4 & 56.5 & 43.3 & 69.2 \\
\bottomrule
\end{tabular}
}

\subcaptionbox{Basketball SideView}{
\begin{tabular}{lccccc}
\toprule
Method & HOTA & DetA & AssA & MOTA & IDF1 \\
\midrule
BoT-SORT & 75.2 & 79.2 & 71.4 & 94.3 & 85.9 \\
ByteTrack & 76.2 & 75.5 & 76.9 & 89.3 & 88.6 \\
\bottomrule
\end{tabular}
}

\subcaptionbox{Basketball SideView2}{
\begin{tabular}{lccccc}
\toprule
Methods & HOTA & DetA & AssA & MOTA & IDF1 \\
\midrule
BoT-SORT & 47.3 & 67.6 & 33.1 & 80.2 & 50.8 \\
ByteTrack & 42.9 & 54.7 & 33.7 & 65.0 & 53.6 \\
\bottomrule
\end{tabular}
}
\subcaptionbox{Basketball TopView}{
\begin{tabular}{lccccc}
\toprule
Method & HOTA & DetA & AssA & MOTA & IDF1 \\
\midrule
BoT-SORT & 66.3 & 62.7 & 70.3 & 89.0 & 93.9 \\
ByteTrack & 65.7 & 65.1 & 66.4 & 89.6 & 92.0 \\
\bottomrule
\end{tabular}
}

\subcaptionbox{Handball SideView}{
\begin{tabular}{lccccc}
\toprule
Methods & HOTA & DetA & AssA & MOTA & IDF1 \\
\midrule
BoT-SORT & 75.1 & 75.5 & 74.7 & 91.6 & 89.7 \\
ByteTrack & 73.5 & 73.8 & 73.2 & 89.4 & 87.6 \\
\bottomrule
\end{tabular}
}

\label{tab:tracking_results}
\end{table}

\begin{table}[h]
\centering
\caption{ByteTrack tracking metrics compared across various datasets, including average scores from TeamTrack and individual scores from sports with the highest (Basketball-SV) and lowest (Basketball-SV2) HOTA scores.}
\scalebox{0.95}{
\begin{tabular}{lccccc}
\toprule
Dataset & HOTA & DetA & AssA & MOTA & IDF1 \\
\midrule
TeamTrack Ave. & 61.9 & 64.2 & 60.2 & 77.2 & 77.5 \\
Basketball-SV & 76.2 & 75.5 & 76.9 & 89.3 & 88.6 \\
Basketball-SV2 & 42.9 & 54.7 & 33.7 & 65 & 53.6 \\
MOT17 & 63.1 & 64.5 & 62 & 80.3 & 77.3 \\
DanceTrack & 47.1 & 70.5 & 31.5 & 88.2 & 51.9 \\
\bottomrule
\end{tabular}
}
\label{tab:tracking_metrics_over_datasets}
\end{table}

\section{Conclusion}
In this study, we introduced TeamTrack, a dataset for multi-object tracking (MOT). The TeamTrack dataset captures object appearances and movements across football, basketball, and handball games using full-pitch, high-resolution videos. Our work included experiments in object detection, trajectory forecasting, and MOT. Our findings reveal opportunities for further research. Specifically, the TeamTrack dataset has limitations regarding team and venue diversity. Expanding the dataset to include a wider range of teams and locations could improve the model generalization. We hope the TeamTrack dataset will contribute to the development of more effective tracking models.

\vspace{-5pt}
\section{Acknowledgements}
This work was financially supported by JSTSPRING Grant-Number JPMJSP2125 and JST PRESTO JPMJPR20CA.
{
    \small
    \bibliographystyle{ieeenat_fullname}
    \bibliography{main}
}

\appendix
\section{Change Log}
\myparagraph{Version 1 (2024-04-22)}

- Initial arXiv version.


\end{document}